\documentclass[sigconf]{acmart}
\AtBeginDocument{%
  }

\copyrightyear{2026}
\acmYear{2026}
\setcopyright{cc}
\setcctype{by}
\acmConference[KDD '26]{Proceedings of the 32nd ACM SIGKDD Conference on Knowledge Discovery and Data Mining V.2}{August 09--13, 2026}{Jeju Island, Republic of Korea}
\acmBooktitle{Proceedings of the 32nd ACM SIGKDD Conference on Knowledge Discovery and Data Mining V.2 (KDD '26), August 09--13, 2026, Jeju Island, Republic of Korea}
\acmDOI{10.1145/3770855.3818983}
\acmISBN{979-8-4007-2259-2/2026/08}
\usepackage{listings}
\usepackage{multirow}
\usepackage{booktabs}

\usepackage{colortbl}
\usepackage{etoolbox}
\usepackage{tabularx}
\usepackage{subcaption}
\usepackage{array}
\usepackage{tcolorbox}

\newcommand{\rulesep}{\unskip\ \vrule\ }

\newcolumntype{C}{>{\centering\arraybackslash}X}

\begin{document}

\title{An Infectious Disease Spread Simulation Based on Large Language Model Decision Making}

\author{Yonchanok Khaokaew}
\authornotemark[2]
\affiliation{
 \department{Computer Science and Engineering\\ Faculty of Engineering}
  \institution{The University of New South Wales}
  \city{Sydney}\state{NSW}
  \country{Australia}
}
\email{y.khaokaew@unsw.edu.au}

\author{Ruochen Kong}
\affiliation{%
  \department{Department of Computer Science}
  \institution{Emory University}
  \city{Atlanta}\state{Georgia}
  \country{USA}
}
\email{ruochen.kong@emory.edu}

\author{Andreas Z\"ufle}
\authornotemark[1]
\affiliation{%
   \department{Department of Computer Science}
  \institution{Emory University}
  \city{Atlanta}\state{Georgia}
  \country{USA}
}
\email{azufle@emory.edu}
\author{Hao Xue}
\affiliation{%
  \institution{The Hong Kong University of Science and Technology (Guangzhou)}
  \city{Guangzhou}
  \country{China}
}
\email{haoxue@hkust-gz.edu.cn}
\author{Taylor Anderson}
\affiliation{%
  \department{Department of Geography and Geoinformation Science}
  \institution{George Mason University}
  \city{Fairfax}\state{Virginia}
  \country{USA}
}
    \email{tander6@gmu.edu}
\author{C. Raina MacIntyre}
\affiliation{%
    \department{The Kirby Institute \\ Faculty of Medicine \& Health}
  \institution{The University of New South Wales}
  \city{Sydney}\state{NSW}
  \country{Australia}
}
\email{r.macintyre@unsw.edu.au}
\author{Matthew Scotch}
\affiliation{%
\department{College of Health Solutions}
  \institution{Arizona State University}
  \city{Tempe}\state{Arizona}
  \country{USA}
}
\email{matthew.scotch@asu.edu}
\author{Flora D. Salim}
\authornotemark[1]

\affiliation{%
  \department{Computer Science and Engineering \\ Faculty of Engineering}
  \institution{The University of New South Wales}
  \city{Sydney}\state{NSW}
  \country{Australia}
}
\email{flora.salim@unsw.edu.au}
\author{David J. Heslop}
\authornotemark[1]
\affiliation{%
  \department{School of Population Health\\ Faculty of Medicine \& Health}
  \institution{The University of New South Wales}
  \city{Sydney}\state{NSW}
  \country{Australia}
}
\email{d.heslop@unsw.edu.au}

\thanks{\raggedright* Corresponding authors: flora.salim@unsw.edu.au, azufle@emory.edu, d.heslop @unsw.edu.au} \thanks{ $\dagger$ Also with King Mongkut's University of Technology North Bangkok (KMUTNB)}

\renewcommand{\shortauthors}{Khaokaew et al.}
\newcommand{\hlnewtwo}[1]{\textcolor{blue}{#1}}

\def\andreas#1{{\small{\sc\textcolor{red}{Andi says: #1}}}}
\def\prooo#1{{\small{\sc\textcolor{blue}{Pro says: #1}}}}

\begin{abstract}

Modelling individual decision-making during infectious disease outbreaks is crucial for understanding behavioural dynamics and informing effective public health interventions. Prior work has shown that large language models can simulate realistic human behaviour by generating agent decisions based on demographic prompts and situational context. We build on this foundation with a spatially grounded, agent-based simulation framework that integrates LLM-generated decisions about self-reported influenza-like illness into a census-based synthetic population of agents. Location is treated as a central feature: agents are assigned to spatial units within cities, capturing the spatial distributions of different demographic groups using real-world census data and enabling geographically diverse behavioural modelling. We implement and compare three decision scenarios, independent reasoning, household influence, and message framing, and simulate self-reporting outcomes in San Francisco and Atlanta. Results reveal that income and education are the dominant drivers of reporting rate variation, with smaller but consistent effects from geography, LLM model choice, and message framing. Our framework generates synthetic data that captures both social and geographic heterogeneity, supporting spatial epidemiological modelling and bias-aware behavioural analysis.

\end{abstract}



\received{10 February 2026}
\received[revised]{18 May 2026}
\received[accepted]{7 June 2026}

\maketitle
\newcommand\kddavailabilityurl{https://github.com/cruiseresearchgroup/disease-simulator-LLM_agent}
\ifdefempty{\kddavailabilityurl}{}{
\begingroup\small\noindent\raggedright\textbf{Resource Availability:}\\
The simulation source code is available at \url{\kddavailabilityurl}.
\endgroup
}

\section{INTRODUCTION}
Simulating human behaviour in-silico offers a safe and privacy-preserving way to explore decision-making in critical scenarios such as pandemics~\cite{zufle2024silico}. Recent advances in large language models (LLMs) have led to the creation of generative agents that mimic human reasoning and behaviour~\cite{acharya2025agentic,park2023generative,xi2025rise,shanahan2023role}. These agents can be embedded in rich simulation environments, following daily life patterns and interacting with one another. Such simulacra hold promise as testbeds for studying population-level responses and designing public health strategies.

Consider the following example: a person wakes up with mild influenza-like symptoms on a weekday morning. They live in a densely populated urban area, enjoy visiting cat cafés, and have recently seen public health advisories about the rising number of cases. Faced with the decision of whether to wear a mask or cancel their plans, their choice may depend on a mix of personal habits, perceived risk, and social responsibility. Typically, our understanding of health behaviours and decision-making comes from individual-level surveys. These surveys can be used to design and parameterise models that simulate human decision-making under different scenarios from which disease outcomes emerge \cite{von2023framework, kong2024infectious}. However, surveys are costly and time-consuming. Instead, to simulate decision-making, we hypothesise that we could prompt LLMs with the individual’s demographic background and contextual information. The LLM then generates a behavioural response, enabling large-scale simulations of nuanced, everyday health decisions.

In prior work, we introduced a behaviour-driven agent-based simulation for modelling individual actions during disease outbreaks, incorporating demographic attributes, activity routines, and probabilistic disease transmission. A logistic regression model trained on individual-level survey data was used to simulate variation in reporting across demographic groups~\cite{kong2024infectious,kong2025simulated}. This framework was built upon a general simulation engine for daily human behaviour patterns~\cite{zufle2023urban}.
%
In this paper, we extend that line of work by replacing the logistic regression with LLM-based decision-making. For each agent, decisions such as whether to report illness or accept a vaccine are generated using an LLM conditioned on the agent’s demographic profile and situational context. This transition enables demographically sensitive and contextually varied behaviours during decision generation. Our simulation results reveal systematic differences in predicted decisions across various demographic factors, illustrating how LLM-driven agents can capture behavioural heterogeneity and support the study of population-level dynamics in disease simulations.  To balance realism with scalability, we pre-generate decisions using several open-source LLMs and store them in a structured decision bank indexed by demographic combinations. During simulation, agents retrieve decisions from this bank rather than querying models in real time. This approach allows us to test the effects of different models, prompt formulations, and contextual framings on simulated behaviour, while ensuring experimental consistency and reproducibility.

By combining structured simulations with the expressive reasoning capabilities of LLMs, we explore the potential of generative agents to mirror survey-based health behaviour data. Our goal is to evaluate whether these agents faithfully reproduce observed behavioural trends or introduce unintended behaviour profiles, and to assess their suitability as behavioural proxies for public health research. Our main contributions are as follows: 1) We introduce an LLM-driven behavioural decision framework integrated into an agent-based disease simulation model, replacing prior rule-based mechanisms. 2) We simulate three real-world-inspired behavioural scenarios: independent decisions, family-shared influence, and public health message framing.
3) We evaluate four open-source LLMs across varying prompt styles and contextual richness, highlighting variability and behaviour profile in model outputs and demonstrating that LLM-generated decisions reflect real-world demographic disparities.

\begin{figure*}[h!]
\centering
\includegraphics[width=0.71\textwidth]{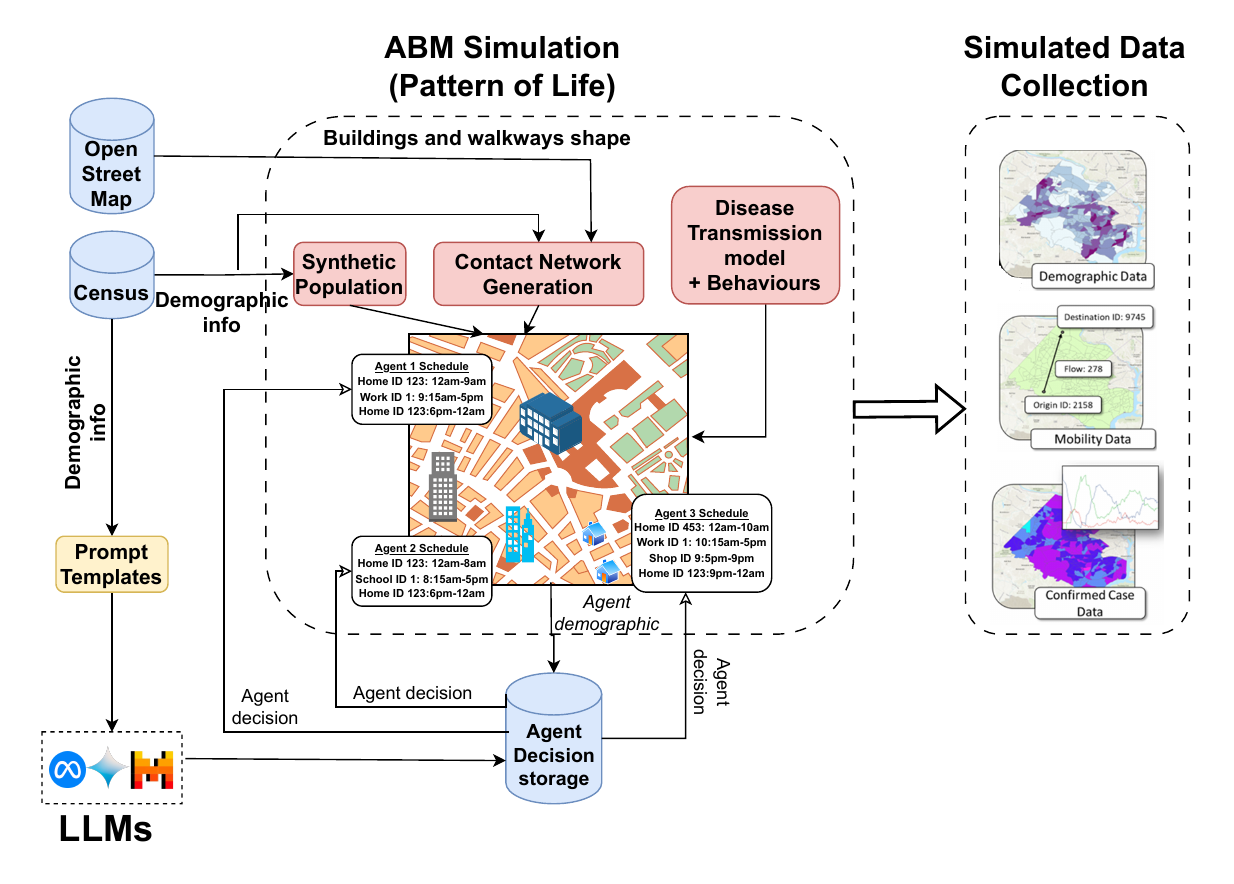}
\vspace{-0.7cm}
\caption{ Infectious Disease Data Simulator: Overview.  }
\Description{ Infectious Disease Data Simulator: Overview.  }
\label{fig:thrust2}
\end{figure*}
\section{RELATED WORKS}\label{sec:related-work}

Most closely related to the problem of predicting the spread of infectious diseases is spatiotemporal prediction, which models variables or events across space and time. This area has been widely explored in domains such as road traffic~\cite{xu2015mining,gkountouna2020traffic,snowdon2018spatiotemporal} and human mobility, including bike sharing and public transport~\cite{liu2019deeppf,lin2017real,islam2021spatiotemporal}. These models often incorporate auxiliary datasets, such as weather conditions or infrastructure layouts, and benefit from large-scale data that accurately reflects real-world patterns. For example, traffic flow data, even when incomplete, remains reliable because shared environmental constraints affect all vehicles.

In contrast, infectious disease data are heavily influenced by behavioural factors, e.g., reporting decisions during the pandemic. The decision to report symptoms or seek testing is shaped by personal circumstances and systemic inequities. For instance, a construction worker without paid leave may be less likely to report an illness, while a university professor with remote work options may be more likely to report one. Numerous factors have been shown to affect reporting behaviour, including occupation~\cite{tostmann2020strong}, symptom recognition~\cite{boelle2020excess}, ethnicity~\cite{dodds2020covid}, frailty~\cite{henwood2020care}, place of residence~\cite{elarde2021change}, social connectedness~\cite{kuchler2020geographic}, internet access~\cite{antoun2016comparisons}, and even an individual's willingness to engage with research~\cite{tyrrell2021genetic}. As a study by~\citet{griffith2020collider} highlights, these reporting biases can significantly distort population-level health measures.

To study such complexities, agent-based models (ABMs) provide a flexible simulation framework. ABMs represent individual agents with diverse characteristics and decision rules, enabling the study of emergent population-level behaviour. While many disease-focused ABMs~\cite{anderson2020neat,pesavento2020data,muscatello2017translation,9162189} successfully model disease transmission, they often assume full observability of infections and lack mechanisms for simulating selective reporting. Recent work has begun to incorporate reporting bias into ABMs, enabling the generation of synthetic datasets that reflect how demographic and behavioural traits influence observed disease outcomes compared to true infections~\cite{kong2025simulated,zufle2024leveraging}. These approaches rely on rule-based mechanisms to simulate decisions like symptom reporting or vaccination. In contrast, we propose a generative approach where agent decisions are produced by large language models prompted with detailed personas and contextual information, including city-specific pandemic scenarios. This allows us to simulate reporting behaviour that is demographically sensitive and contextually varied, capturing nuanced individual and spatial differences.

Our work contributes to the emerging class of generative agent-based simulations, which use LLMs to model agent cognition and behaviour. Prior studies have explored LLM-driven agents in domains such as social interaction~\cite{park2023generative,shanahan2023role}, policy making~\cite{xiao2023simulating,qian2024scaling}, software development~\cite{qian-etal-2024-chatdev}, and healthcare~\cite{williams2023epidemic}. In the health domain, \citet{williams2023epidemic} used ChatGPT to simulate individual movement and isolation decisions in a town-wide epidemic. However, prior healthcare simulations did not examine reporting disparities or incorporate spatial demographic variation. Our study extends this by focusing on symptom reporting, integrating multiple LLMs, and analysing how demographic and geographic context influence modelled behaviour.

\section{SIMULATION FRAMEWORK}\label{sec:framework}
We build upon the open-source Patterns of Life simulator~\cite{zufle2023urban}, a Java-based platform that models human behaviour through Maslowian needs~\cite{maslow1943theory}. In this framework, agents pursue daily goals that satisfy their core needs: returning home fulfils the Shelter Need, eating (at home or in restaurants) addresses the Food Need, going to work satisfies Financial Needs, and engaging with others meets the Love Need. Agent behaviour is guided by the Theory of Planned Behaviour~\cite{ajzen1991theory}, with actions planned around individual needs, social factors, and available information. In this paper, each \textit{agent} represents an individual with a distinct demographic profile and daily routine, capable of interacting with the environment and others

We extend this simulation environment in three key ways. First, we incorporate an infectious disease transmission model based on SEIR dynamics, allowing agents to progress through susceptible, exposed, infectious, and recovered states according to physical interactions. Second, we initialise the synthetic population using attributes sampled from real-world census data, providing a realistic demographic foundation for each agent. Third, we replace the prior regression-based decision mechanism with LLM-based decision-making. Instead of manually defining behaviour rules, we generate agent decisions (e.g., whether to report symptoms) using LLMs prompted with demographic and contextual information. These extensions enable the simulation of more diverse, realistic, and demographically profiled behaviours, as illustrated in Figure~\ref{fig:thrust2}. They also enable us to assess whether LLM-generated decisions replicate or worsen behavioural disparities observed in real populations. 
\subsection{Infectious Disease Model}




We simulate disease transmission using an extended SEIR model~\cite{kermack1932contributions}. SEIR is appropriate for influenza-like illness because the Exposed state captures the defined incubation period and the Recovered state captures temporary immunity; more complex compartmental alternatives would introduce additional parameters without adding explanatory value for our primary contribution, which is the behavioural decision layer. Each agent begins in the \textbf{Susceptible (S)} state and may become infected through contact with nearby \textbf{Infectious (I)} agents in shared physical locations. Once infected, the agent transitions to the \textbf{Exposed (E)} state, where they are not yet infectious. After an incubation period of $d_E$ simulation days, the agent becomes \textbf{Infectious (I)} and can spread the disease to others with probability $p_I$.

To reflect variation in symptom presentation, we introduce a \textbf{Symptomatic} sub-state within the Infectious state. Upon entering this state, agents are probabilistically classified as either symptomatic or asymptomatic. Only symptomatic agents undergo changes in behaviour and reporting decisions. They remain at home for $d_{\text{home}}$ days and initiate a reporting process driven by LLMs.

\begin{figure}[t]
\centering
\includegraphics[width=\linewidth]{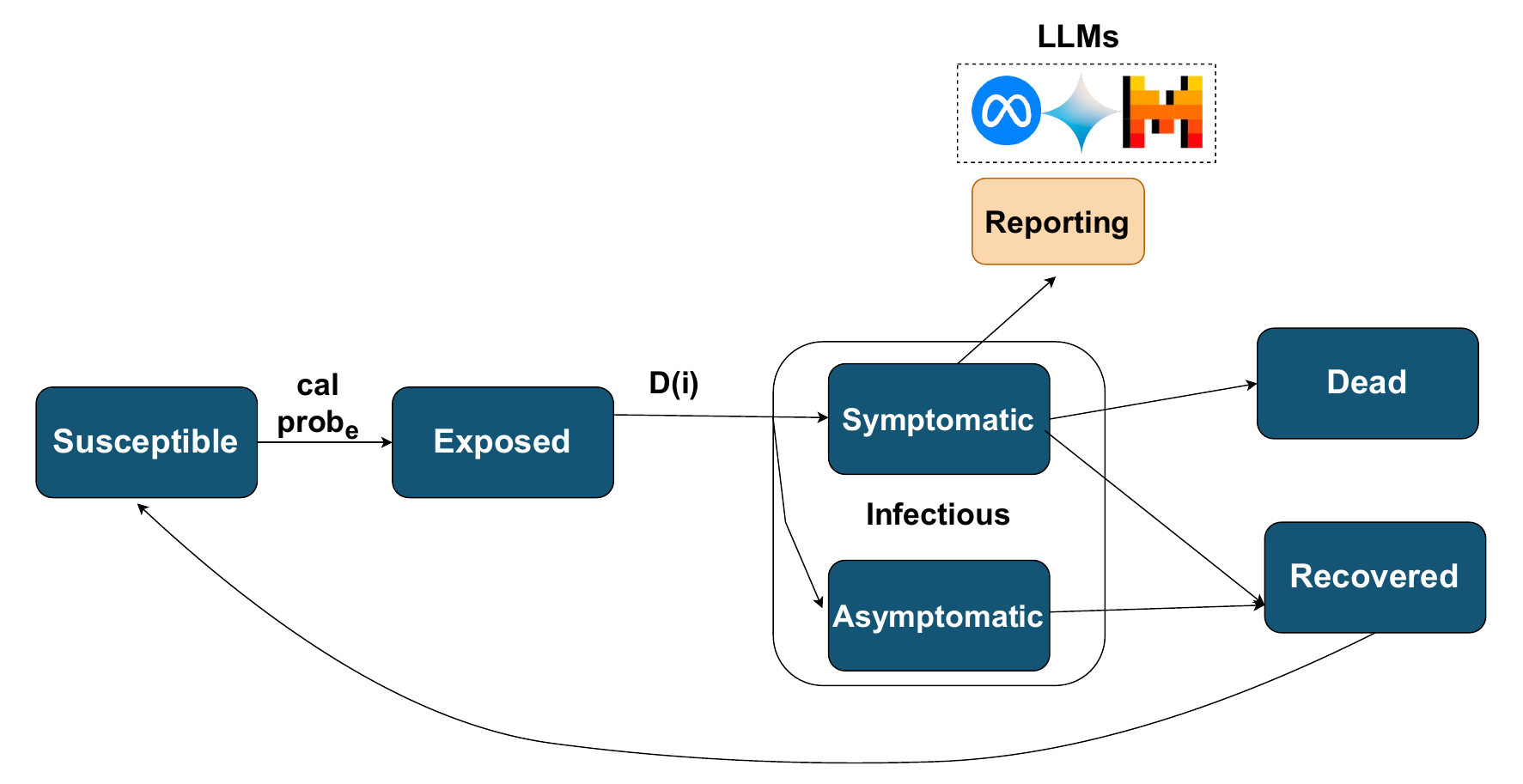}
\caption{Infectious Disease Model, including infection (exposure), disease progression, and final outcomes.}
\Description{Infectious Disease Model, including infection (exposure), disease progression, and final outcomes.}
\label{fig:SIR}
\end{figure}

When symptomatic, the agent queries its pre-generated LLM decision record to determine whether it should report the symptoms. If the LLM response is “No”, the agent makes no behaviour change beyond home isolation, and the case remains unreported. If the LLM response is “Yes”, the agent proceeds to a reporting pathway with two probabilistic stages: they test positive with probability $p = 0.05$, or receive a clinical diagnosis with probability $p = 0.5$ if the test is negative or unavailable ~\cite{tokars2018seasonal, ma2018healthcare}. In either case, the agent is recorded as a reported case in the public health system and is placed in isolation. If neither occurs, the case remains unreported. This extended structure enables us to simulate realistic reporting dynamics, combining individual-level decisions with diagnostic uncertainty, and allows us to analyse how behavioural and structural factors influence the observed number of cases in surveillance data.

\begin{figure*}[!h]
\rulesep\captionsetup[subfigure]{aboveskip=-1pt,belowskip=-1pt}
\centering

\begin{subfigure}{0.17\linewidth}
\centering
\includegraphics[width=\linewidth,height=2cm]{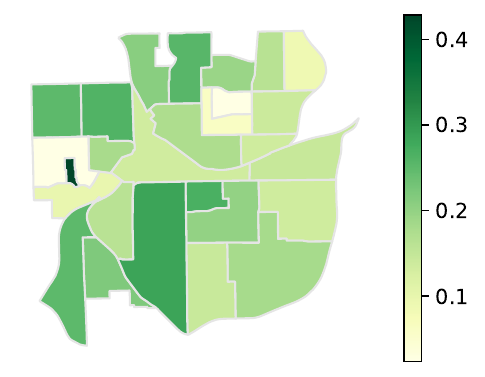}
\caption{Age>50 Years}
\label{fig:censusage-atl}
\end{subfigure}\rulesep
\begin{subfigure}{0.17\linewidth}
\centering
\includegraphics[width=\linewidth,height=2cm]{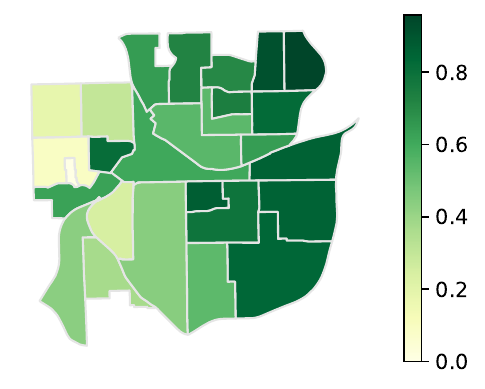}
\caption{Income>\$50,000}
\label{fig:censusincome-atl}
\end{subfigure}\rulesep
\begin{subfigure}{0.17\linewidth}
\centering
\includegraphics[width=\linewidth,height=2cm]{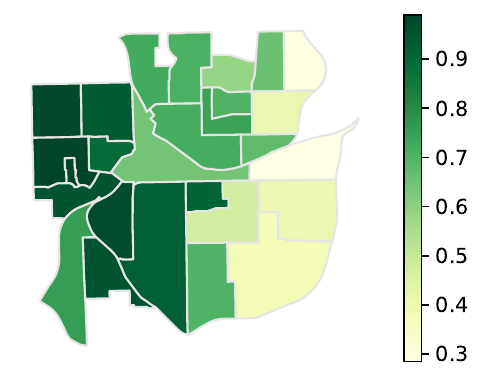}
\caption{Race not White}
\label{fig:censusrace-atl}
\end{subfigure}\rulesep
\begin{subfigure}{0.17\linewidth}
\centering
\includegraphics[width=\linewidth,height=2cm]{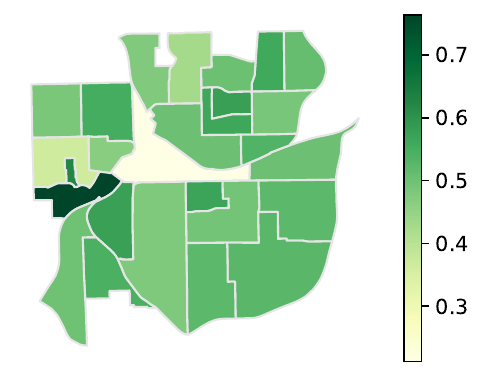}
\caption{Gender Female}
\label{fig:censusgender-atl}
\end{subfigure}\rulesep
\begin{subfigure}{0.17\linewidth}
\centering
\rulesep
\includegraphics[width=\linewidth,height=2cm]{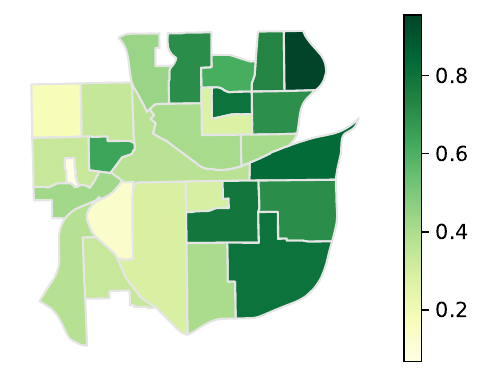}
\caption{Bachelors or higher}
\label{fig:censusedu-atl}
\end{subfigure}\rulesep
\begin{subfigure}{0.17\linewidth}
\centering
\includegraphics[width=\linewidth,height=2cm]{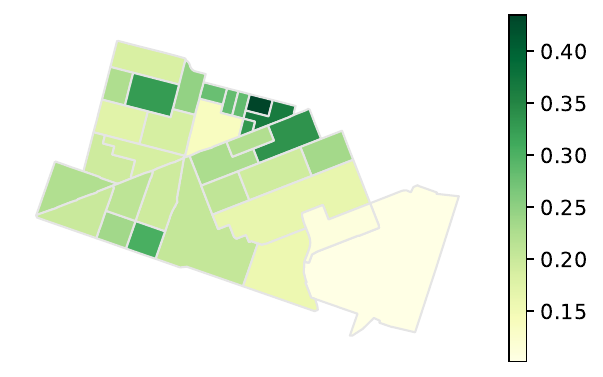}
\caption{Age>50 Years}
\label{fig:censusage-sf}
\end{subfigure}\rulesep
\begin{subfigure}{0.17\linewidth}
\centering
\includegraphics[width=\linewidth,height=2cm]{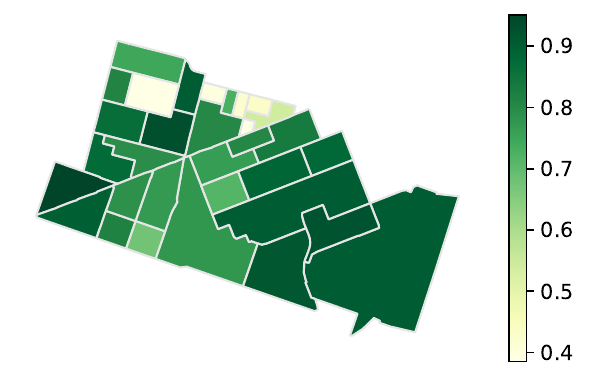}
\caption{Income>\$50,000}
\label{fig:censusincome-sf}
\end{subfigure}\rulesep
\begin{subfigure}{0.17\linewidth}
\centering
\includegraphics[width=\linewidth,height=2cm]{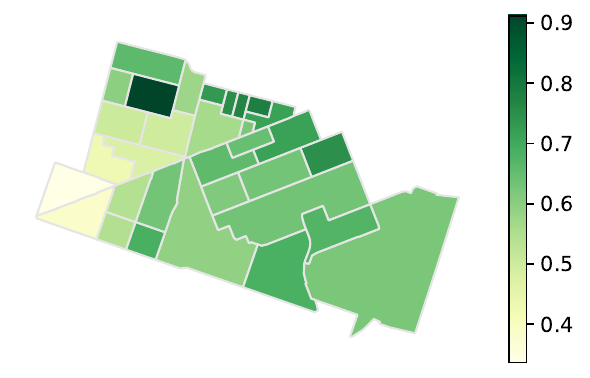}
\caption{Race not White}
\label{fig:censusrace-sf}
\end{subfigure}\rulesep
\begin{subfigure}{0.17\linewidth}
\centering
\includegraphics[width=\linewidth,height=2cm]{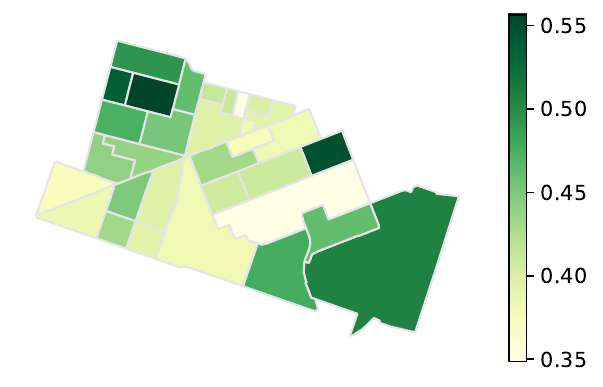}
\caption{Gender Female}
\label{fig:censusgender-sf}
\end{subfigure}\rulesep
\begin{subfigure}{0.17\linewidth}
\centering
\includegraphics[width=\linewidth,height=2cm]{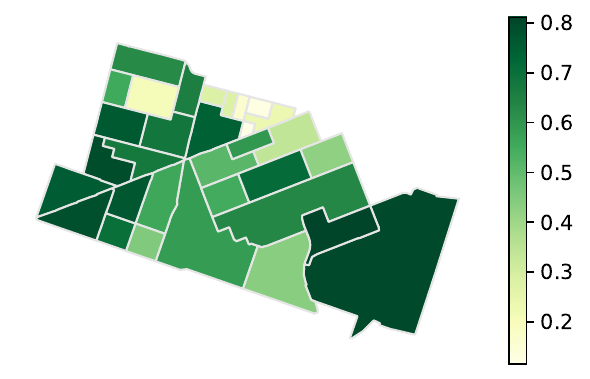}
\caption{Bachelors or higher}
\label{fig:censusedu-sf}
\end{subfigure}\rulesep

%
%
\caption[-]{Socioeconomic Census Data for Atlanta (Top) and San Francisco  (Bottom) }
\Description{Socioeconomic Census Data for Atlanta and San Francisco, showing demographic distributions.}

\label{fig:census}

\end{figure*}
\subsection{Agent Generation with Census Data}

To ensure that our synthetic population reflects realistic demographic and geographic distributions, we generate agents using real-world census data rather than uniform random sampling. The simulation map is divided into census regions, specifically, census tracts for the United States, using publicly available demographic and boundary datasets\footnote{https://www2.census.gov/geo/tiger/TIGER2020PL/STATE/, https://www.census.gov/geographies/mapping-files/2020/geo/tiger-line-file.html}.  Agent synthetic populations are then generated using conditional probabilities based on the actual population of each tract, so that more densely populated regions in the real world are proportionally represented in the simulation. Within each tract, agent attributes such as age, gender, race, education, and income are drawn according to local census distributions. This method accounts for demographic variation at the neighbourhood level and preserves spatial disparities. Our system is flexible: additional attributes can be included by extending the source files with relevant census variables.

For this study, we applied this generation procedure to downtown areas of San Francisco and Atlanta. Figure~\ref{fig:census} illustrates the resulting spatial and demographic patterns. Subfigures~\ref{fig:census}a–e show San Francisco, and~\ref{fig:census}f–j show Atlanta. In Atlanta, age and gender distributions are relatively uniform across tracts (Figures~\ref{fig:censusage-atl} and~\ref{fig:censusgender-atl}), but income and racial composition (Figures~\ref{fig:censusincome-atl} and~\ref{fig:censusrace-atl}) vary markedly from east to west, with higher income and White populations concentrated in western tracts. In contrast, San Francisco shows less pronounced variation across tracts. Some edge cases, such as the north of San Francisco or the west of Atlanta, show unusual attribute distributions, which can be explained by low population counts and higher variance in those census tracts. This demographic grounding supports more realistic and spatially aware agent behaviour and enables the study of how demographic inequalities align with reporting bias and disease visibility in the simulation.

\subsection{LLM-Based Agent Decision Making}

\begin{figure}[!ht]
\centering
\includegraphics[width=0.9\linewidth]{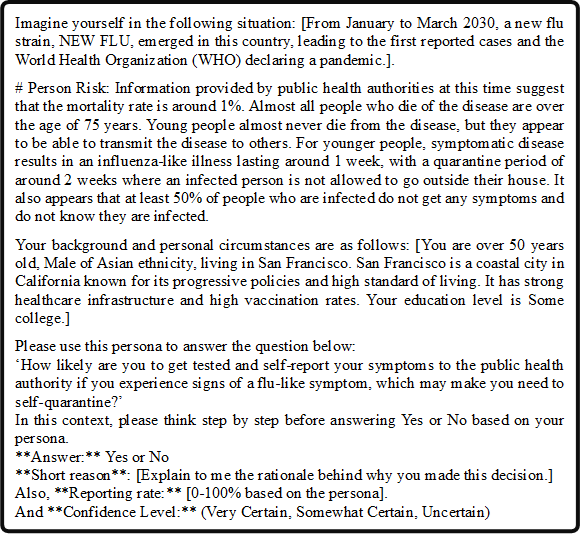}
\vspace{-0.1cm}
\caption{Prompt example}
\Description{Prompt example}
\vspace{-0.7cm}
\label{fig:prompt_ex}
\end{figure}

To simulate realistic symptom reporting behaviour, we pre-generate decisions using LLMs for all possible combinations of demographic attributes. These combinations are created by crossing binary or categorical values of five key features: age (under or over 50), race (White, Black, Asian, or Other), gender (male or female), education (high school or below, some college, bachelor’s or above), and income (above or below \$70,000).  This feature set, aligned with previous synthetic population frameworks ~\cite{von2023framework, 
kong2025simulated}, ensures compatibility and allows for direct comparison of behavioural differences. By using consistent input variables, variations in LLM outputs (e.g., stronger income effects than those of logistic models) can be attributed to generative reasoning. Additionally, these features align with established epidemiological findings, which identify socioeconomic status and race as key factors influencing testing compliance and health-seeking behaviour~\cite{mody2021understanding, zhu2021association}, confirming their importance in realistic disease modelling. This results in a set of hypothetical profiles, including combinations that may not appear in the real population. Each profile is encoded as a five-digit key. Scenarios that require additional contextual input extend this to a six-digit key, where the extra digit encodes the relevant context dimension (such as household influence or message framing), routing each agent to the appropriate conditional bank at runtime. Pre-generating decisions for all key combinations allows the simulation to scale without real-time LLM inference. For each key, we query the LLM at least five times using a consistent prompt and record the number of “Yes” and “No” responses (see Fig.~\ref{fig:prompt_ex} for an example prompt). For instance, a row in a structured decision bank might look like: \texttt{|10121 | 4 | 1|}, indicating that four out of five responses suggested the agent would report symptoms, and one response did not.

At runtime, when an agent becomes symptomatic and reaches the nightly reporting stage, it uses its demographic key to retrieve the corresponding decision pool and randomly selects one response. If the result is “Yes”, the agent reports; if “No”, it continues its routine. For the main simulation, we construct a combined decision bank by aggregating responses from four open-source LLMs. This approach provides agents with a diverse yet consistent behavioural foundation without relying on any single model. To analyse model-specific variation, we also generate separate decision banks for each LLM. These are used in our experiments to compare how different models interpret the same prompts and demographic profiles. While LLMs aren’t designed to replicate human health behaviour, research indicates they don't consistently produce ideal or socially optimal responses. They may reveal hesitancy or conflicting decisions, making them intriguing for modelling realistic variation. LLMs are increasingly examined in healthcare simulation, policy-making, and scenario planning fields that benefit from insights into behavioural variability. We aim to investigate whether LLM-based decision generation can serve as a scalable and adaptable proxy for simulating public health behaviour.

\subsubsection{Simulation Scenarios}
To evaluate how agent decisions are influenced by demographic, social, and informational factors, we implemented three experimental scenarios (detailed in Appendix~\ref{appendix:prompt}). Each varies the context encoded in the demographic key; scenarios requiring additional contextual input use the six-digit extension described above to route agents to the appropriate conditional bank at runtime. These scenarios allow us to assess behavioural consistency, model sensitivity, and the impact of social or informational context on disease reporting.

\textit{Scenario 1 (Independent):} In this baseline scenario, agents make decisions independently based solely on their demographic profiles and a fixed prompt. Before the simulation begins, each agent’s decision is pre-generated using LLMs. The LLM prompt includes demographic attributes such as age, gender, income, education, and race, along with a standard situational context (e.g., experiencing flu symptoms).

\textit{Scenario 2 (Household Influence):} Agents are provided with a household-aware context: prompts explicitly inform the agent that a family member has reported an illness, and a separate pre-computed bank is generated for this condition. When a symptomatic agent detects a reporting household member at runtime, a sixth digit is appended to the demographic key, routing the query to the family-aware bank instead of the baseline pool. This design is extensible: additional key digits could encode neighbourhood-level incidence bins, temporal conditions, or other contextual signals, enabling finer-grained contextual adaptation without real-time LLM inference.

\textit{Scenario 3 (Message Framing):} tests how public health messaging affects reporting. Agents are randomly assigned one of three framings: Risk-Based (personal health consequences), Altruism-Based (protection of others), or Data-Based (statistical evidence). Separate LLM decision banks are generated for each framing, enabling comparison of how message type influences overall reporting rates and equity across demographic groups.

\begin{table*}[!ht]
\resizebox{0.92\linewidth}{!}{\begin{tabular}{|r|r|r|l|l|l|l|l|}
\hline
\multicolumn{1}{|c|}{\textbf{step}} & \multicolumn{1}{c|}{\textbf{agentId}} & \multicolumn{1}{c|}{\textbf{regionId}} & \multicolumn{1}{c|}{\textbf{diseaseStatus}} & \multicolumn{1}{c|}{\textbf{diseaseSeq}} & \multicolumn{1}{c|}{\textbf{time}} & \multicolumn{1}{c|}{\textbf{location}} & \multicolumn{1}{c|}{\textbf{checkin}} \\ \hline
7255                                & 1352                                  & 8                                      & Infectious                                  & 1709-1.1274-1.1352-1                     & 2019-07-26T04:35:00.000            & POINT (...)                            & AtHome                                \\ \hline
7270                                & 4723                                  & 26                                     & Recovered                                   & \multicolumn{1}{c|}{-}                   & 2019-07-26T05:50:00.000            & POINT (...)                            & AtHome                                \\ \hline
7272                                & 3255                                  & 18                                     & Recovered                                   & \multicolumn{1}{c|}{-}                   & 2019-07-26T06:00:00.000            & POINT (...)                            & AtHome                                \\ \hline
7275                                & 3325                                  & 19                                     & Infectious (Symptomatic)                                 & 2040-1.3685-1.3325-1                     & 2019-07-26T06:15:00.000            & POINT (...)                            & AtHome                                \\ \hline
7281                                & 1289                                  & 8                                      & Recovered                                   & \multicolumn{1}{c|}{-}                   & 2019-07-26T06:45:00.000            & POINT (...)                            & AtHome                                \\ \hline
7282                                & 3160                                  & 17                                     & Exposed                                     & ?.3160-1                                 & 2019-07-26T06:50:00.000            & POINT (...)                            & AtRecreation                          \\ \hline
7283                                & 63                                    & 1                                      & Infectious                                  & 3-1.46-1.63-1                            & 2019-07-26T06:55:00.000            & POINT (...)                            & AtHome                                \\ \hline
7285                                & 4386                                  & 23                                     & Infectious (Symptomatic)                                       & ?.4386-1                                 & 2019-07-26T07:05:00.000            & POINT (...)                            & AtHome                                \\ \hline
7293                                & 3625                                  & 20                                     & Exposed                                     & ?.3625-1                                 & 2019-07-26T07:45:00.000            & POINT (...)                            & AtWork                                \\ \hline
\bottomrule
\end{tabular}}
\caption{Example output of the generated infectious disease case data, recording each agent's disease-status change event over time and space.\vspace{-0.7cm}}

\label{tabel:disease-output-sample}
\end{table*}

\subsection{Simulation Setting}
Our experiments utilised four open-source large language models (LLMs): Meta Llama-3-8B-Instruct~\cite{dubey2024llama}, Google Gemma-2-9B-IT~\cite{team2024gemma}, Mistral AI Mistral-8B-Instruct~\cite{jiang2023mistral}, and Galactica-6.7B-Evol-Instruct~\cite{taylor2022galactica}. All models were accessed via the Hugging Face platform, and the appropriate licenses were obtained prior to experimentation. To ensure consistency across models, we set the generation parameters with a temperature of 0.6 and a top-p value of 0.9. 


For the infectious disease simulation, we adopted an extended SEIR model. The infection probability was set to $p_I = 0.07$ (a fixed per-contact transmission probability that does not change during the simulation), with the infectious period $d_I$ sampled from 5 to 8 days and the recovery duration $d_R$ sampled from 30 to 180 days to represent temporary immunity. The exposure duration $d_E$, representing the incubation period, was sampled over the range 1 to 5 days. Symptomatic probability was estimated using age-dependent estimates from the Covasim model~\cite{kerr2020covasim}, in which the probability of developing symptoms and the likelihood of a more acute case increase with age. Only symptomatic agents were eligible to report symptoms and self-isolate. All of these parameters are adjustable within the simulation environment, allowing researchers to adapt the disease dynamics to different pathogens or outbreak scenarios.

\section{SIMULATION RESULT}\label{sec:results}

\subsection{Simulation Scenario Outcomes}\label{sec:results_out}

\begin{figure}[!h]
\centering
\begin{subfigure}[c]{0.49\linewidth}
\centering
\includegraphics[width=\linewidth]{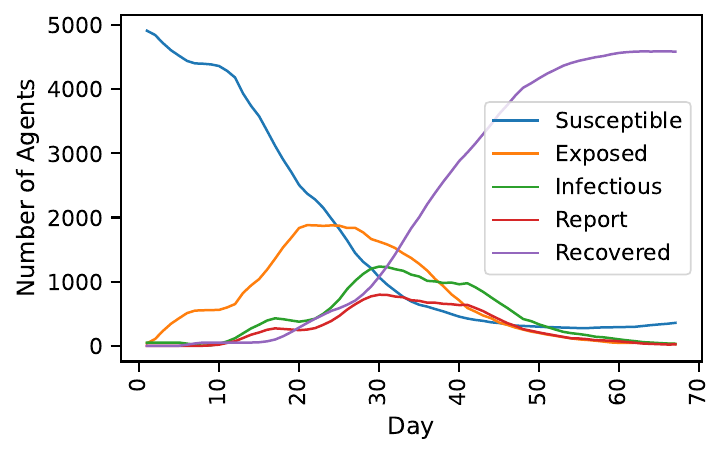}
\caption{Atlanta }
\label{fig:epi1}
\end{subfigure}
\begin{subfigure}[c]{0.49\linewidth}
\centering
\includegraphics[width=\linewidth]{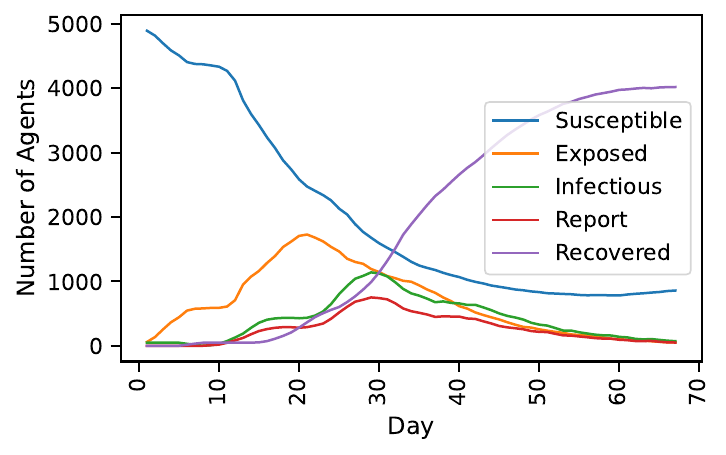}
\caption{San Francisco  }
\label{fig:epi2}
\end{subfigure}

\caption{Epidemic curve for infectious diseases spread in the two areas ( $d_E=7-14, p_I=0.07,d_I=5-8, d_R=30-180$) }
\Description{Epidemic curve}
\vspace{-4px}
\label{fig:epi}
\end{figure}

Before evaluating the impact of different simulation scenarios, we first present an example of the disease progression output generated by our simulator. 
Table~\ref{tabel:disease-output-sample} illustrates how infections are tracked over time and space, recording each agent’s disease-status change with its simulation step, timestamp, check-in location, and home region ID. This structure supports downstream spatiotemporal analyses such as transmission mapping or hotspot detection. The diseaseSeq column traces the transmission chain in X-Y format, where X-Y denotes the $Y$-th infection of Agent $X$; for example, \texttt{1709-1.1274-1.1352-1} denotes a path from Agent 1709 to 1274 to 1352. A ``?’’ indicates that the infecting agent did not report, reflecting observability gaps caused by selective reporting. Epidemic curves (Figure~\ref{fig:epi}) follow expected SEIR dynamics, while reported case variation reflects demographic-driven reporting differences.

\begin{figure}[!h]
\centering

\begin{subfigure}{\linewidth}
\centering
\includegraphics[width=\linewidth]{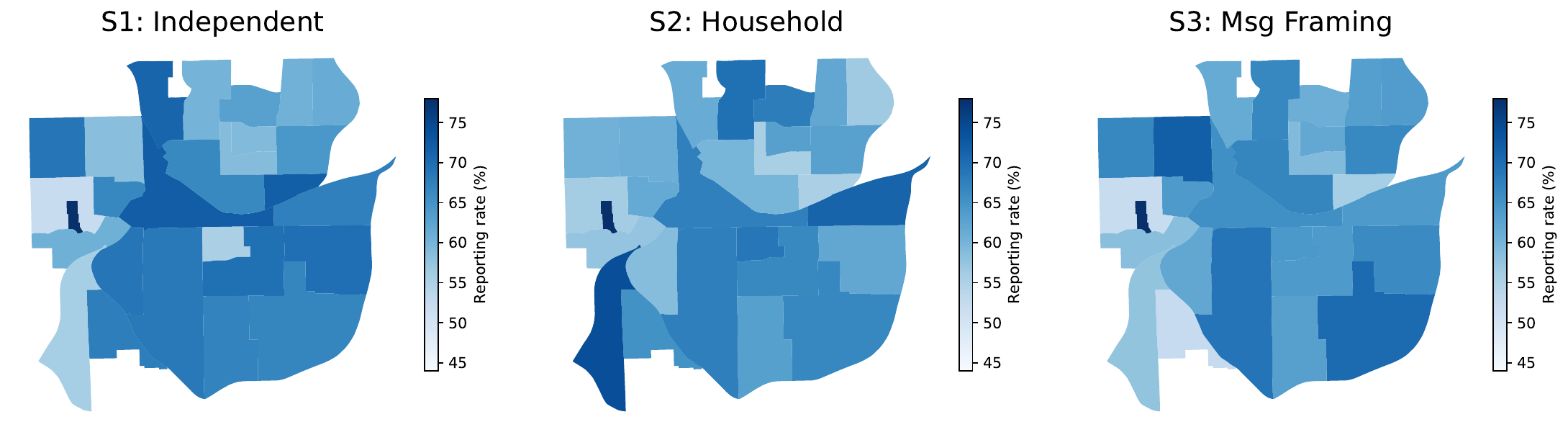}
\caption{Atlanta}
\label{fig:scALT1}
\end{subfigure}
\begin{subfigure}{\linewidth}
\centering
\includegraphics[width=\linewidth]{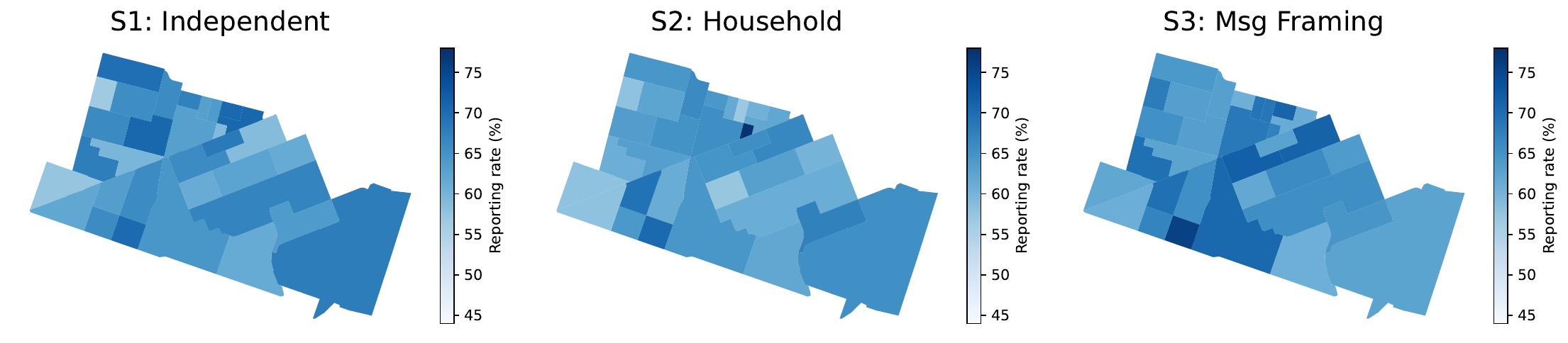}
\caption{San Francisco}
\label{fig:scSAN1}
\end{subfigure}

\caption{Reporting rate on different scenarios}
\Description{Reporting rate on different scenarios}
\vspace{-8px}
\label{fig:rerate}
\end{figure}

To assess the impact of demographic, social, and informational factors on symptom reporting, we evaluated all three simulation scenarios across both Atlanta and San Francisco (Figure~\ref{fig:rerate}). Under the baseline independent scenario (S1), mean reporting rates were 65.4\% in Atlanta and 64.7\% in San Francisco. Reporting behaviour aligned closely with demographic disparities: census tracts with higher income and education levels (Figure~\ref{fig:census}b, e, g, and j) exhibited elevated reporting rates, particularly in the eastern tracts of Atlanta and the southern districts of San Francisco, while lower reporting was observed in areas with a higher proportion of non-white populations (Figure~\ref{fig:census}c and h), such as western Atlanta and some part of the north region of San Francisco.

Introducing household-level influence in Scenario~2 modestly reduced mean reporting in both cities (Atlanta: 64.0\%, San Francisco: 63.5\%) and minimised the spread of rates across tracts in Atlanta (standard deviation 6.4\% compared to 7.6\% in S1). This suggests that intra-household conformity, while capable of reinforcing reporting when family members are already active, can also pull moderate-reporting agents downward when household hesitancy is prevalent.

Scenario~3 (message framing) produced more differentiated effects. In San Francisco, mean reporting rose to 65.9\%, and the lowest-performing tracts improved by approximately 4 percentage points, indicating that varied informational framings particularly benefited previously low-engagement areas. In Atlanta, the mean was 63.9\%, with a standard deviation similar to S1, suggesting that message framing can lift specific tracts but does not uniformly shift reporting in a city with stronger underlying demographic stratification. Overall, demographic attributes remain strong predictors of reporting behaviour under neutral prompts, while message framing showed a positive effect on equity, particularly in San Francisco.

\subsection{Comparing against Logistic Regression}
To quantify the specific value-add of generative agents, we benchmark our LLM-driven simulation against the prior logistic regression (LR) baseline established in \citet{kong2025simulated}. While the LR model serves as a robust standard for predicting health behaviours based on demographic features, it is fundamentally constrained by the linearity of its decision boundary and the assumption of feature independence.
Table \ref{tab:comparison} summarises the qualitative and quantitative divergence between the two approaches. The limitation of the LR baseline is its tendency to regress to the mean; as shown in prior work, the LR model typically predicts reporting probabilities clustered around 0.5 (range: 0.38-0.96). In contrast, the LLM agents exhibit a full spectrum of behavioural responses (0.04-1.00), effectively capturing the long tail of non-compliant or highly anxious subpopulations.

\begin{table}[h]
\centering
\small
\caption{Comparative analysis of the agent decision-making between the Logistic Regression baseline }
\vspace{-5px}
\resizebox{\columnwidth}{!}{%
\begin{tabular}{@{}p{1.3cm}p{3.8cm}p{4.5cm}@{}}
\toprule
\textbf{Metric} & \textbf{Logistic Baseline}  \cite{kong2024infectious} & \textbf{Generative LLM Agent (Ours)} \\ \midrule
\textbf{Behavioural \newline Diversity} & \textbf{Constrained:} Probabilities clustered near mean (0.38-0.96). & \textbf{High Variance:} Full spectrum of probabilities (0.04-1.00), capturing outliers. \\ \midrule
\textbf{Income \newline Sensitivity} & \textbf{Weak:} Minimal impact on reporting probability. & \textbf{Moderate-Strong:} Explicitly models economic barriers to reporting. \\ \midrule
\textbf{Age \newline Reasoning} & High reporting correlates strictly with older age (>65). &  Captures reporting incentives for the workforce (19-50) to protect employment/peers. \\ 

\bottomrule
\end{tabular}%
}
\vspace{-10px}

\label{tab:comparison}
\end{table}

Another observation is that the logistic regression model assumes independent variables are mutually independent (or requires manual interaction terms). However, demographic features such as \textit{Income} and \textit{Education} are highly correlated in real-world census data. The LR baseline tends to overestimate the independent effect of high income while dampening the nuance of intersectionality. In contrast, the LLM agents appear to implicitly model these correlations, generating decisions that reflect the compounded pressure of low income and low education without requiring manual feature engineering. Furthermore, Logistic regression can struggle with highly imbalanced classes (rare reporting events). We observe that the LLM is less prone to smoothing these rare behaviours toward the mean, successfully capturing specific subgroups (e.g., low-income workers) who systematically avoid reporting due to economic constraints.

To quantify directional agreement, we computed Spearman rank correlations between LLM-predicted reporting rates and LR-predicted rates across 71 matched demographic groups. We find $\rho = 0.416$ ($p = 0.013$) in Atlanta and $\rho = 0.411$ ($p = 0.013$) in San Francisco. Income and education ordering is preserved in both cities, confirming that the LLM agents reproduce the key demographic gradients of the logistic baseline despite operating via a fundamentally different mechanism.

As a second independent proxy, we compare LLM reporting rates against COVID-19 vaccine intent from the Understanding America Study~\cite{kapteyn2024understanding}. We use vaccine intent as a directional proxy because both vaccination and symptom reporting reflect community-oriented health-seeking behaviour shaped by the same socioeconomic barriers. On income, reporting rises from low to high in both datasets (74\% to 96\% simulated; 54\% to 81\% UAS), and education shows the same gradient (68\% to 97\% simulated; 54\% to 76\% UAS). Race ordering is also consistent across both (Asian $>$ White $>$ Black). Disagreements on gender and age are expected: COVID vaccine intent is strongly age-risk-driven, whereas ILI self-reporting follows a different age-based dynamic. These findings suggest that we should present our agents as behavioural proxies, which show consistent validity across key demographic dimensions.

\subsection{LLM Variation Analysis}

In this section, we investigate how variation in LLMs, prompt formulation, and contextual richness influences simulated agent reporting behaviour. These experiments help us understand how different modelling choices affect the final simulation outputs and whether such variation introduces spatial or demographic bias.

\begin{figure*}[!h]
\centering
\begin{subfigure}{0.3\linewidth}
\centering
\includegraphics[width=\linewidth]{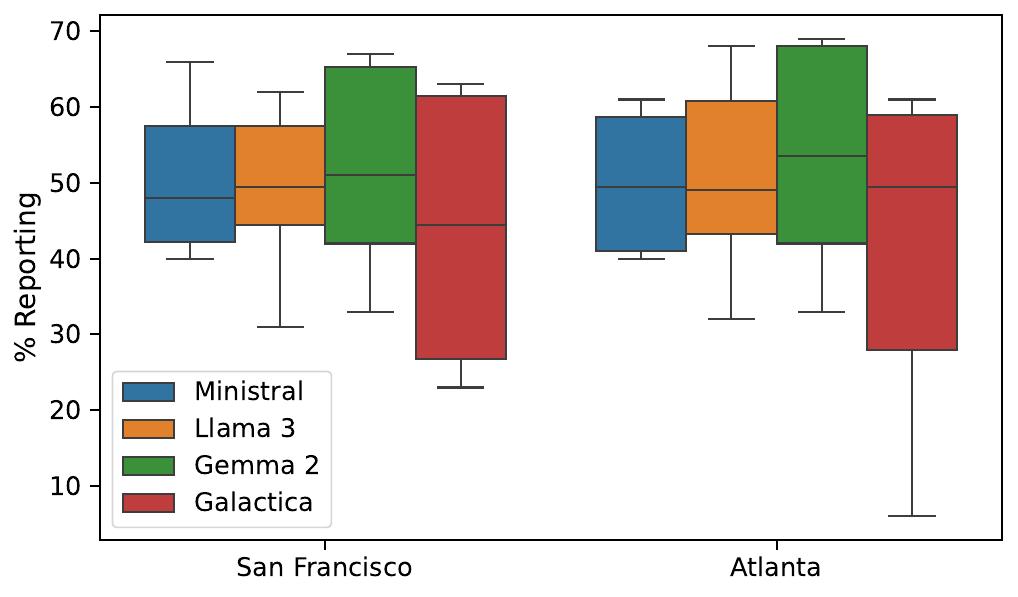}
\caption{Different models and cities }
\label{fig:llm-variation}
\end{subfigure}
\begin{subfigure}{0.3\linewidth}
\centering
\includegraphics[width=\linewidth]{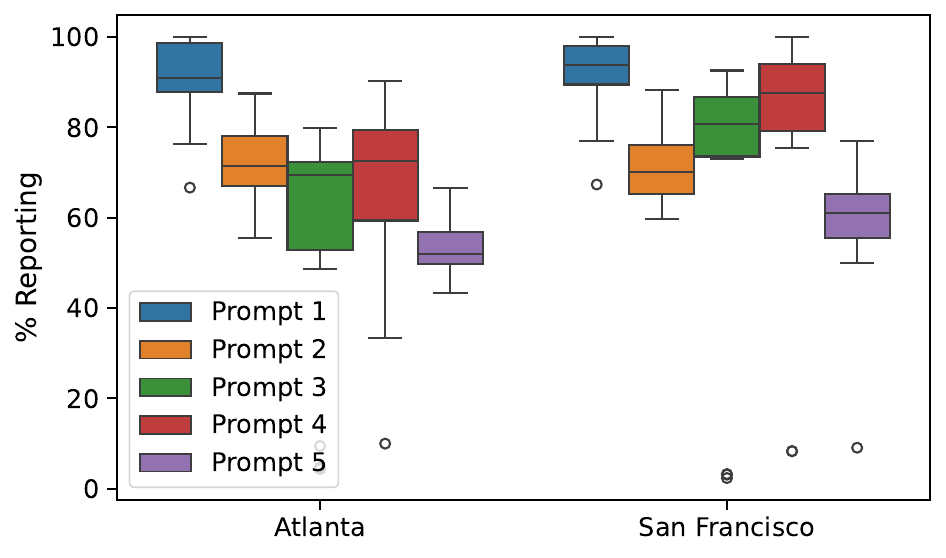}
\caption{Different prompt formulations}
\label{fig:llm-prompt-sensitivity}
\end{subfigure}
\begin{subfigure}{0.3\linewidth}
\centering
\includegraphics[width=\linewidth]{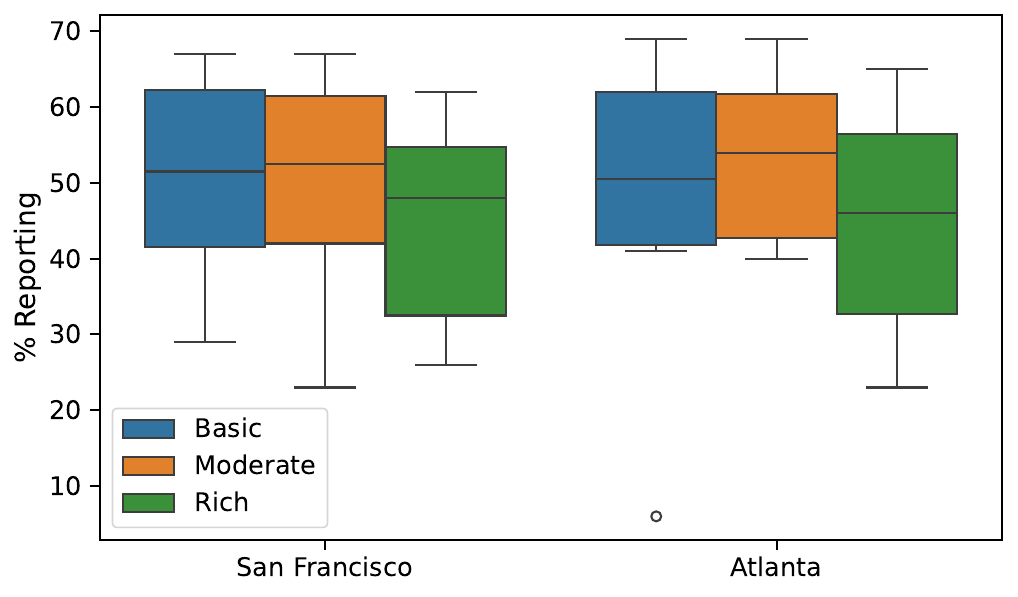}
\caption{Varying levels of contextual richness}

\label{fig:compare richcontext}
\end{subfigure}
\vspace{-2px}
\caption{Comparison of LLM-predicted reporting behaviour across different settings}
\Description{Comparison of LLM-predicted reporting behaviour across different settings}
\vspace{-7px}
\end{figure*}

We focus on three aspects: (a) differences across LLMs when used to generate agent decisions; (b) sensitivity to changes in prompt wording and framing; and (c) the impact of providing richer geographic and social context in the prompt. Each experiment is based on either the outputs from LLMs or downstream simulation results, as explained in each subsection.
\subsubsection{Model Variation by City}

We examined how the choice of LLM affects agent decision-making by generating a separate decision bank from each model, holding all demographic inputs and prompt structure constant. Figure~\ref{fig:llm-variation} shows the distribution of predicted reporting rates from each model across demographic profiles and cities, using a prompt exploration dataset. The median rates across models range from approximately 40--65\%, indicating that LLMs tend to predict symptom reporting as a behaviour that is neither universally adopted nor completely avoided. Note that the main simulation decision bank, which includes more contextual information, yields higher overall rates (depending on the demographic group); Section~\ref{sec:results_out} reports these simulation-level values.

Additionally, agents in Atlanta reported a slightly higher rate than those in San Francisco for some models, most notably Gemma~2. Although the same prompts were used for both cities, this pattern was consistent across models. This suggests that geographic context may influence LLM outputs, possibly due to implicit associations about healthcare access, cultural attitudes, or socioeconomic factors. These findings highlight geography as a potential feature worth further investigation in LLM-driven behavioural simulations.

We included the Galactica model to assess whether scientific-domain pretraining produces different behavioural patterns than general-purpose instruction models. A post-hoc ablation comparing the full 4-model ensemble to a 3-model ensemble excluding Galactica yields Pearson $r = 0.993$ ($p < 10^{-135}$) with a mean absolute rate difference of 2.23\%. So, Galactica does not drive our findings; its inclusion tests a deliberate model-diversity hypothesis and confirms that scientific text pretraining alone is insufficient to alter demographic decision patterns on this task.

\subsubsection{Prompt Sensitivity}

We evaluated prompt sensitivity by testing five variants that differed not merely in wording but in the contextual information presented to the agent, including personal risk, decision rationale, and family decision (Figure~\ref{fig:llm-prompt-sensitivity}). Prompts 1, which asked a straightforward question about symptom reporting given basic demographics, produced the highest rates (near 90\%). Prompts 2 and 3 reframed the decision by asking for the reason behind it and outlining a sequence of testing, self-reporting, and possible quarantine, leading to a sharp decline in Atlanta. In contrast, San Francisco declined slightly because agents facing an explicit quarantine cost naturally reduce their reported willingness. Prompt 4 restored average rates to 75-85\% in both cities by adding mortality, personal risk and transmission statistics that activated health-seeking reasoning. Prompt 5, which anchored the decision to a household member's prior choice, produced moderate and stable rates of 55--65\%. These shifts occur because agents respond to contextual framing rather than superficial wording, altering city-level reporting rates by 25–35\% from the baseline. Despite this absolute variation, demographic rank ordering remains stable across all five variants, preserving the direction and magnitude of income and education effects.

\subsubsection{Impact of Context detail}

Contextual richness introduces a design trade-off (Figure~\ref{fig:compare richcontext}). Basic prompts produced substantial inter-model variability in both cities, with similarly wide core distributions across models. Enriched context narrowed median rates but triggered distinct outlier behaviours, most notably a conservative low outlier in Atlanta under the richest context, while overall prediction ranges remained broader in Atlanta than in San Francisco. While richer context enhances realism, this resulting model divergence is a critical consideration for policy-oriented applications (see Appendix~\ref{sec:contextual}). The canonical prompt for our main results uses the rich context level (city-specific pandemic scenario; see Appendix~\ref{appendix:prompttest}). Here, the main demographic effect estimates have a median 95\% CI of $\pm$3.6 pp (Atlanta) and $\pm$1.7 pp (San Francisco). Despite this contextual variation, demographic attributes remained the strongest and most consistent driver of reporting decisions.

\subsection{Impact of Demographics on Reporting}

\begin{figure*}[!h]
\centering
\begin{subfigure}{0.195\linewidth}
\centering
\includegraphics[width=\linewidth]{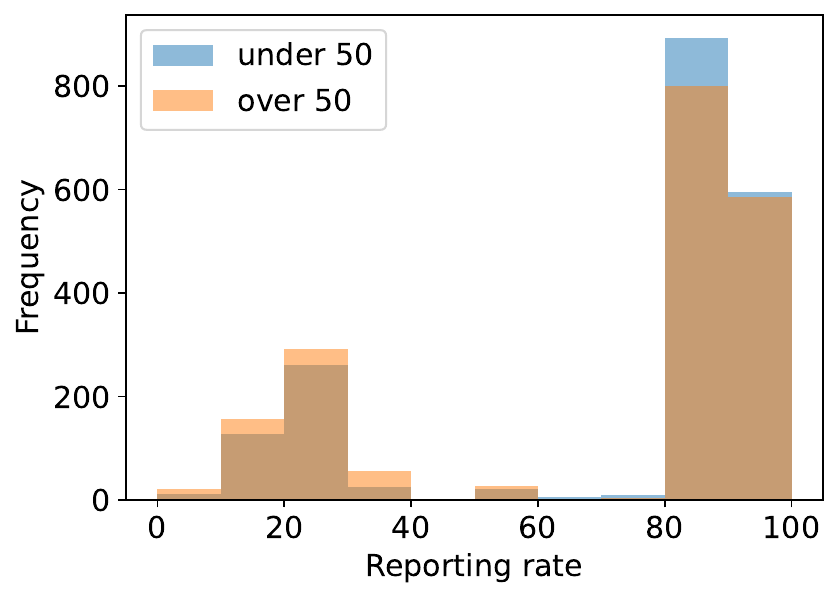}\vspace{-0.0cm}
\caption{Age \vspace{-0.0cm}}
\label{fig:biasage}
\end{subfigure}
\centering
\begin{subfigure}{0.195\linewidth}
\centering
\includegraphics[width=\linewidth]{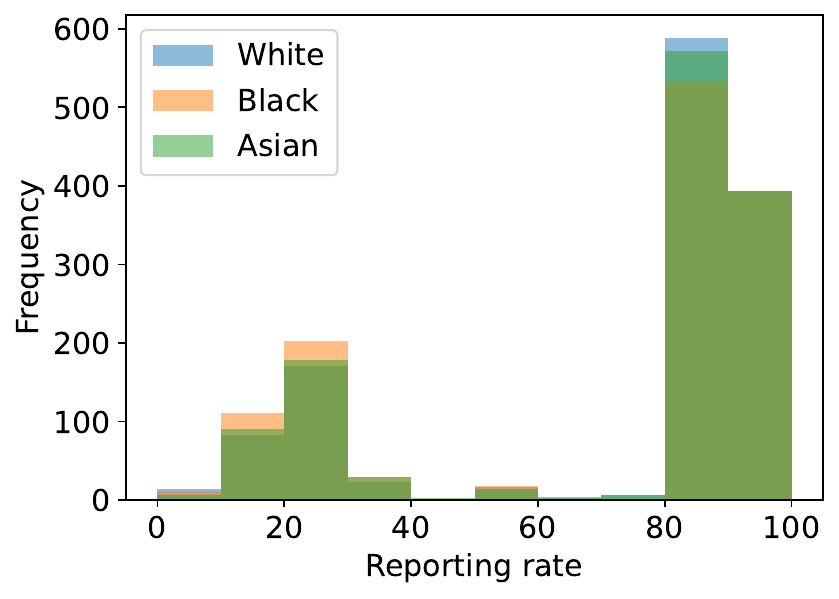}\vspace{-0.0cm}
\caption{Race \vspace{-0.0cm}}
\label{fig:biasrrac}
\end{subfigure}
\centering
\begin{subfigure}{0.195\linewidth}
\centering
\includegraphics[width=\linewidth]{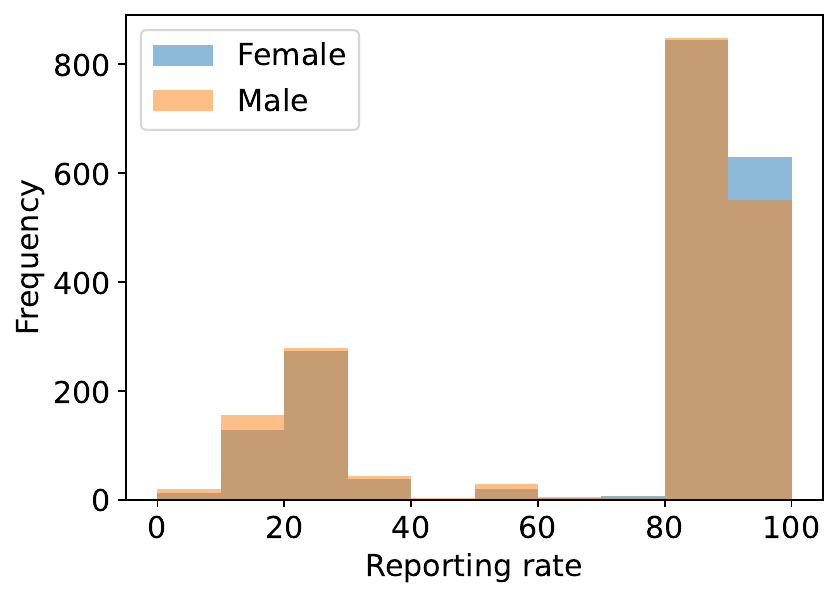}\vspace{-0.0cm}
\caption{Gender\vspace{-0.0cm}}
\label{fig:biasgen}
\end{subfigure}
\begin{subfigure}{0.195\linewidth}
\centering
\includegraphics[width=\linewidth]{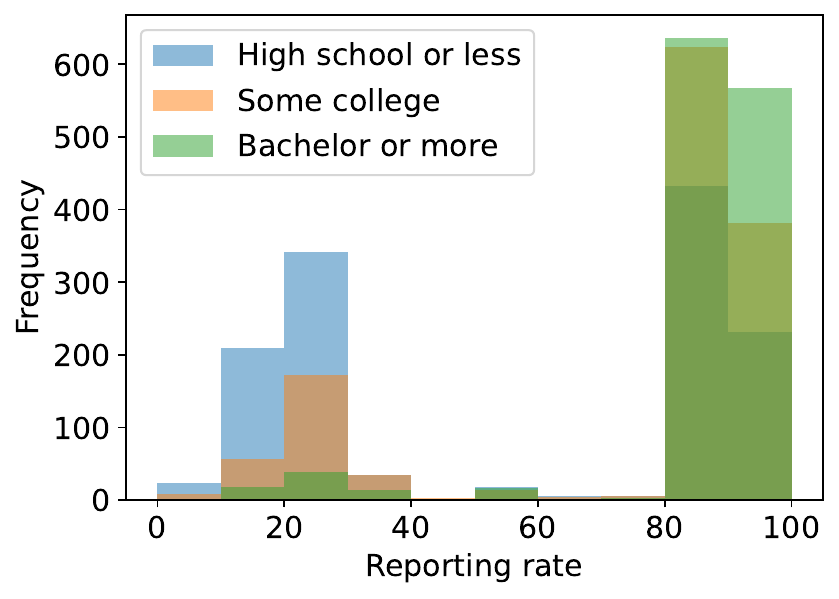}\vspace{-0.0cm}
\caption{Education\vspace{-0.0cm}}
\label{fig:biaseed}
\end{subfigure}
\begin{subfigure}{0.195\linewidth}
\centering
\includegraphics[width=\linewidth]{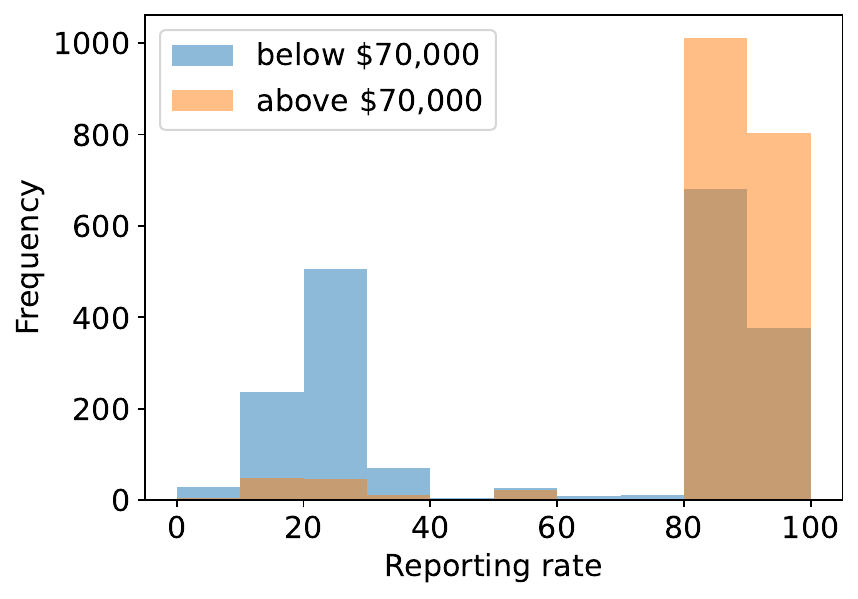}\vspace{-0.0cm}
\caption{Income\vspace{-0.0cm}}
\label{fig:fig:biasin}
\end{subfigure}
\vspace{-2px}
\caption{Distribution of predicted reporting rates across individual demographic attributes.}
\Description{Distribution of predicted reporting rates across individual demographic attributes.}
\vspace{-6px}
\label{fig:reporting-distribution}
\end{figure*}

We further analysed the distributions of reporting-rate predictions across demographic profiles. This allowed us to move beyond binary outcomes and examine the reporting rate as a continuous variable, providing more nuanced insights into variations within and across demographic groups. Figure~\ref{fig:reporting-distribution} shows the distribution of predicted reporting rates across five individual demographic variables: income, education, race, gender, and age. The most noticeable differences appear in income and education. Agents with high income (above \$70k) or a bachelor’s degree consistently received higher predicted reporting rates, often clustered around 80–95\%. In contrast, agents with low income or only a high school education saw lower, more dispersed values. Race, gender, and age showed minor individual effects but still drove overall variation.

To quantify these patterns, we applied one-way ANOVA, linear regression, and post-hoc pairwise comparisons (Tukey's HSD). Table~\ref{tab:anova} summarises the results. Income and education had the largest effects on predictions ($\eta^2 = 0.1972$ and $0.1675$, respectively). Because the decision bank uses a balanced factorial design that crosses all demographic combinations, these variables are orthogonal (VIF = 1.00), preventing multicollinearity. The LLM model and city showed smaller effects ($\eta^2 = 0.0436$ and $0.0370$), while age, race, and gender had very small effects. These patterns align with real-world data: prior studies found lower-income households were significantly less likely to test and report symptoms during the COVID-19 pandemic~\cite{zhu2021association, mody2021understanding}, validating LLM-driven simulations for modelling public health engagement.

These findings show that LLM-generated decisions are not uniform across the population. By reflecting differences based on income, education, and geography, the model outputs reveal potential disparities in simulated public health behaviours. However, we also observe small but consistent differences between LLMs, even when presented with the same prompts and contexts. This highlights the importance of model selection: not all LLMs behave equally, and their individual characteristics can significantly influence simulation outcomes in subtle yet meaningful ways. Therefore, validation of model choice, ideally against real-world behavioural data, is crucial when using LLMs for behavioural simulations. These insights support the broader application of LLM-based agents in studying inequality in visibility and compliance, while also cautioning that the choice of model may influence the observations made.

\begin{table}[h!]
\centering

\caption{Statistical analyses on predicted reporting rates across demographic and contextual variables.\vspace{-0.1cm}}
\resizebox{0.85\linewidth}{!}{

\begin{tabular}{|l|r|c|l|c|}
\hline
\textbf{Variable} & \textbf{F-stat.} & \textbf{Effect Size ($\eta^2$)} & \textbf{Effect} & \textbf{p-value} \\
\hline
\hline
Income        & 957.07 & 0.1972 & Large       & $<$0.001 \\
Education     & 391.82 & 0.1675 & Large       & $<$0.001 \\
LLM Model     &  59.15 & 0.0436 & Small       & $<$0.001 \\
City          &  74.79 & 0.0370 & Small       & $<$0.001 \\
Age           &  12.16 & 0.0031 & Very small  & $<$0.001 \\
Race          &   4.00 & 0.0020 & Very small  & 0.018 \\
Gender        &   6.73 & 0.0017 & Very small  & 0.010 \\
\hline

\end{tabular}}
\label{tab:anova}
\end{table}

To confirm that demographic rank ordering is preserved across prompt formulations, we computed 95\% CIs for each group across the five prompt variants (Appendix~\ref{appendix:ci-table}). The high-income group reports at 82.1\% ($\pm$21.0\%) in Atlanta and 86.8\% ($\pm$15.2\%) in San Francisco, consistently above the low-income group (55.8\%\,$\pm$\,14.6\%; 65.0\%\,$\pm$\,19.0\%), and the education gradient (high school $<$ some college $<$ bachelor) holds for every prompt variant in both cities.

These subtle but consistent trends suggest that adding more contextual information does not necessarily lead to higher reporting. One explanation could be that richer prompts introduce greater nuance or ambiguity, which may lead the model to make more conservative or hesitant decisions. This finding highlights an important consideration for simulation design: more information does not always translate into stronger action, especially when working with LLM-based agents that respond sensitively to prompt complexity.

\section{CONCLUSION}\label{sec:conclusion}
This work presents an extension of a spatial agent-based simulation framework that incorporates large language models to generate individual decision-making behaviours during an infectious disease outbreak. By replacing hand-crafted behaviour functions with LLM-driven outputs, we enabled more adaptive and context-sensitive modelling of symptom reporting. Through scenarios that encompass independent decisions, family influence, and message framing, we demonstrated that both social structure and communication strategies significantly affect the visibility of disease cases. Specifically, our experiments showed that LLM-based decision-making reflects underlying demographic disparities, with higher-income and higher-education groups consistently receiving more favourable behavioural outputs. However, interventions such as public health message framing reduced these gaps, suggesting that LLMs can effectively model behaviour-change strategies and produce plausible responses.

Despite these promising results, our findings highlight critical considerations for the design of LLM-driven simulations. First, model selection has a significant impact: different LLMs produced systematically different behaviours even under identical conditions, suggesting that models encode distinct implicit assumptions and reasoning styles. Second, we observed a trade-off between contextual richness and inter-model consistency. While richer context enables more nuanced agent behaviours, it increases variability across models, raising questions about robustness. Consequently, prompt standardisation and uncertainty quantification, such as confidence intervals and scenario-based analyses, are essential when deploying these simulations for policy or public health decisions.

Future work will address these challenges by extending the framework to other decision types, such as vaccination uptake, and incorporating live, context-aware simulations. By integrating retrieval-augmented generation or memory-augmented agents, we aim to capture dynamic, real-time adaptation and improve the alignment between simulated agents and real-world behavioural data.

\section*{LIMITATIONS AND ETHICAL CONSIDERATIONS}
\subsection*{Ethical use of Data and Informed Consent}
This study relies entirely on the generation of synthetic agents to simulate human behaviour, utilising publicly available, aggregated census data to construct demographic profiles. As the simulation operates on behavioural proxies rather than real human subjects, the research did not require an informed consent mechanism typically mandated for human-subjects research. We prioritised privacy by generating all agent attributes, such as income, education, and race, probabilistically from census tract distributions, so that no specific individuals could be identified or re-identified.

However, the use of Large Language Models to proxy for human decision-making raises ethical concerns about algorithmic bias. We acknowledge that LLMs may inadvertently reproduce or amplify stereotypes present in their training data when simulating the behaviours of specific demographic groups. While our findings illustrate that LLM agents can reflect real-world socioeconomic disparities in health reporting, care must be taken to ensure these simulations are used to identify and mitigate structural disadvantages rather than reinforce discriminatory assumptions about compliance or health behaviours in marginalised communities.

\subsection*{Limitations} Our study is subject to several limitations inherent to the current state of generative agents. First, our results highlight a significant dependence on model selection: distinct LLMs (e.g., Llama-3 vs. Mistral) exhibited different baseline reporting rates and sensitivities to demographic cues, even under identical prompting conditions. Second, the agents demonstrated high sensitivity to prompt framing, with minor changes in wording or context leading to substantial shifts in predicted behaviour. For policy-oriented applications, we recommend using simpler, standardised prompts that prioritise reproducibility; experiments with richer contextual detail are best treated as exploratory tools for hypothesis generation rather than direct inputs for policy decisions.

Third, to ensure computational scalability, we utilised a pre-generated decision bank approach. While agents can switch among banks mid-simulation via event keys (as in Scenario 2), the banks themselves are fixed at run time; fully continuous adaptation to arbitrary unfolding conditions, such as arbitrary neighbourhood-level incidence trajectories, is not supported in the current framework, which may oversimplify the complexity of human decision-making during a prolonged pandemic. Finally, our demographic model was limited to five key attributes (age, race, gender, education, and income), excluding other potentially critical factors, such as occupation, household composition, or political affiliation, that likely influence public health compliance. Future work should validate these synthetic behaviours against granular real-world survey data to confirm their predictive accuracy.

\section*{ACKNOWLEDGEMENTS}
This research is supported by the Australian Commonwealth Scientific and Industrial Research Organisation (CSIRO) and the United States National Science Foundation (NSF) under Grant Nos. 2302968, 2302969, and 2302970 (titled "Collaborative Research: NSF-CSIRO: HCC: Small: Understanding Bias in AI Models for the Prediction of Infectious Disease Spread" \cite{zufle2024leveraging}), with additional independent support from the NSF under Grant No. 2109647. We express our gratitude to the NVIDIA Academic Grant Program for providing access to an A100 GPU on Saturn Cloud, and to OpenAI’s Researcher Access Program for API access to GPT models.

\bibliographystyle{ACM-Reference-Format}
\bibliography{main}

\section*{GEN AI DISCLOSURE}
Large Language Models were utilised as a core methodological component of this research to simulate individual decision-making behaviours within the agent-based framework. These models generated the behavioural responses used to analyse reporting trends and disparities across demographic groups. Additionally, generative AI tools were used during manuscript preparation to refine and improve clarity; the authors reviewed all outputs and take full responsibility for the final content.

\appendix
\section{APPENDIX}

\subsection{Prompt Template and Scenarios}\label{appendix:prompt}
\vspace{-9px}
\begin{figure}[!ht]
\centering
\includegraphics[width=\linewidth]{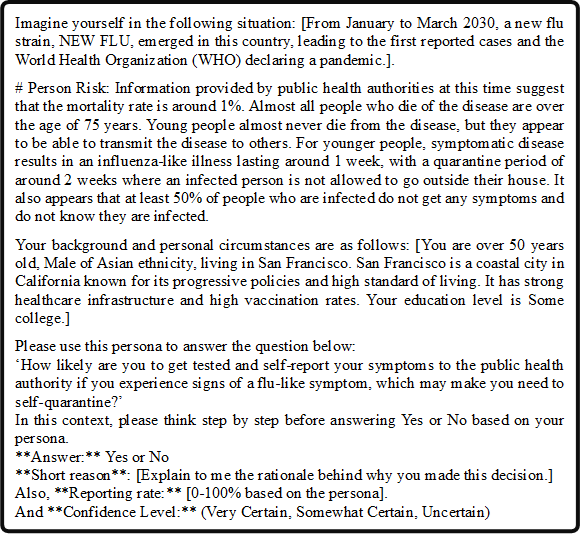}

\caption{Structure of the Baseline System Prompt used in the system}
\Description{Structure of the Baseline System Prompt used in the system}
\vspace{-9px}
\label{fig:prompt_exx}
\end{figure}

Figure~\ref{fig:prompt_exx} illustrates the baseline system prompt structure (Scenario 1). To simulate specific intervention scenarios, we injected additional context into the decision-making phase as follows:

\textbf{Scenario 2: Household Influence}
\begin{tcolorbox}[colback=white!5!white,colframe=black!75!black,boxrule=1pt,arc=.3em,boxsep=-1mm,fontupper=\footnotesize]
In your household, another family member recently experienced similar symptoms and decided to report/not report their illness to public health authorities.
\end{tcolorbox}

\textbf{Scenario 3: Message Framing}

\begin{tcolorbox}[colback=white!5!white,colframe=black!75!black,boxrule=1pt,arc=.3em,boxsep=-1mm,fontupper=\footnotesize]
\textbf{Framing A (Risk-Based):} "Not reporting your symptoms could result in worsening health, delayed treatment, and potential long-term complications."

\textbf{Framing B (Altruism-Based):} "By reporting your symptoms, you help protect your family, co-workers, and community from further spread of illness."

\textbf{Framing C (Data-Based):} "According to recent data, early reporting of symptoms reduces disease transmission by 40\% and increases the chance of recovery with mild symptoms."
\end{tcolorbox}

\subsection{Contextual Prompt}\label{sec:contextual}

This section will detail the varying levels of geographic context provided to the agents during the sensitivity analysis.

\begin{tcolorbox}[colback=white!5!white,colframe=black!75!black,boxrule=1pt,arc=.3em,boxsep=-1mm,fontupper=\footnotesize]
\textbf{Atlanta}

\textbf{Basic:} Atlanta is a major city in the southeastern United States with a diverse population and a mix of urban and suburban areas. Access to healthcare varies across neighbourhoods.

\textbf{Moderate:} Atlanta, Georgia, is a culturally diverse city with notable health disparities between communities. Some neighbourhoods have strong trust in public health systems, while others face barriers due to past systemic inequalities and economic conditions.

\textbf{Rich:} In Atlanta, the population is highly diverse, including large Black and Hispanic communities. While some areas benefit from strong local healthcare infrastructure, others face limited access. Past experiences with unequal healthcare delivery have shaped public attitudes, with mistrust of government communication remaining a barrier in certain areas. Public health messaging during pandemics is often met with mixed responses.
\end{tcolorbox}

\begin{tcolorbox}[colback=white!5!white,colframe=black!75!black,boxrule=1pt,arc=.3em,boxsep=-1mm,fontupper=\footnotesize]
\textbf{San Francisco}

\textbf{Basic:} San Francisco is a coastal city in California known for its progressive policies and high standard of living. It has a strong healthcare infrastructure and high vaccination rates.

\textbf{Moderate:} San Francisco is known for its tech industry and high-income inequality. The city has a relatively health-aware population, but homeless and underserved communities face significant challenges in accessing care and support during public health crises.

\textbf{Rich:} San Francisco’s population is generally responsive to public health guidelines. However, sharp contrasts in income and housing stability lead to unequal access to healthcare. While many residents trust government guidance and technology-driven health tracking, marginalised groups often experience exclusion. Public health officials emphasise community outreach to bridge gaps, particularly in Asian and Latino communities.
\end{tcolorbox}

\subsection{Prompt Sensitivity Analysis}\label{appendix:prompttest}
\begin{tcolorbox}[colback=white!5!white,colframe=black!75!black,boxrule=1pt,arc=.3em,boxsep=-1mm,fontupper=\footnotesize]
\textbf{\textit{Prompt 1}}
\vspace{3pt}

Imagine yourself in the following situation: [From January to March 2030, a new flu strain, NEW FLU, emerged in this country, leading to the first reported cases and the World Health Organisation (WHO) declaring a pandemic.].

\vspace{3px}
Your background and personal circumstances are as follows: 
[You are under [AGE] years old, [GENDER] of [RACE] ethnicity living in [CITY]. [CITY CONTEXT]. 
Your household income is [INCOME]. Your education level is [EDUCATION]].
\vspace{3px}

Please use this persona to answer the question below:

\vspace{3px}

\textit{How likely are you to report your symptoms if you experience signs of a new flu?}

In this context, please think step by step before answering Yes or No based on your persona.

\vspace{3px}
\textbf{Response Format: }

1. Yes or No Answer 

2. Confidence Level: (Very Certain, Somewhat Certain, Uncertain)

\end{tcolorbox}

\begin{tcolorbox}[colback=white!5!white,colframe=black!75!black,boxrule=1pt,arc=.3em,boxsep=-1mm,fontupper=\footnotesize]
\textbf{\textit{Prompt 2}}
\vspace{3px}

Imagine yourself in the following situation: [From January to March 2030, a new flu strain, NEW FLU, emerged in this country, leading to the first reported cases and the World Health Organisation (WHO) declaring a pandemic.].

\vspace{3px}
Your background and personal circumstances are as follows: 
[You are under [AGE] years old, [GENDER] of [RACE] ethnicity living in [CITY]. [CITY CONTEXT]. You're living in a diverse country with varying access to healthcare, differing levels of trust in government and medical institutions, and socioeconomic disparities. 
Your household income is [INCOME]. Your education level is [EDUCATION]].
\vspace{2px}

Please use this persona to answer the question below:
\vspace{2px}

\textit{How likely are you to report your symptoms if you experience signs of a new flu?}

\vspace{3px}

In this context, please think step by step before answering Yes or No based on your persona. 
Answer: [Yes or No]

SHORT REASON:   [Explain to me the rationale behind why you made this] decision. 
And Confidence Level: [(Very Certain, Somewhat Certain, Uncertain)]

\vspace{4px}

\textbf{Response Format: }

1. Yes or No Answer

2. Confidence Level: (Very Certain, Somewhat Certain, Uncertain) 

3. Brief Reason: [one sentence, explain to me the rationale behind why you made this decision.] 
\vspace{4px}

\end{tcolorbox}
\begin{tcolorbox}[colback=white!5!white,colframe=black!75!black,boxrule=1pt,arc=.3em,boxsep=-1mm,fontupper=\footnotesize]
\textbf{\textit{Prompt 3}}
\vspace{3px}

Imagine yourself in the following situation: [From January to March 2030, a new flu strain, NEW FLU, emerged in this country, leading to the first reported cases and the World Health Organisation (WHO) declaring a pandemic.].

\vspace{3px}
Your background and personal circumstances are as follows: 
[You are under [AGE] years old, [GENDER] of [RACE] ethnicity living in [CITY]. [CITY CONTEXT]. You're living in a diverse country with varying access to healthcare, differing levels of trust in government and medical institutions, and socioeconomic disparities. 
 
Your household income is [INCOME]. Your education level is [EDUCATION]].
\vspace{3px}

Please use this persona to answer the question below:

\vspace{3px}

\textit{How likely are you to get tested and self-report your symptoms into the new flu public health system if you experience signs of a new flu, which may make you need to self-quarantine?}

\vspace{3px}

In this context, please think step by step before answering Yes or No based on your persona. 
Answer: [Yes or No]

SHORT REASON:   [Explain to me the rationale behind why you made this] decision. Also, Reporting rate: [0-100\% based on the persona]
And Confidence Level: [(Very Certain, Somewhat Certain, Uncertain)]

\vspace{3px}

\textbf{Response Format: }

1. Yes or No Answer

2. Confidence Level: (Very Certain, Somewhat Certain, Uncertain) 

3. Reporting rate: [0-100\% based on the persona]

4. Brief Reason: [one sentence, explain to me the rationale behind why you made this decision.] 
\vspace{3px}

\end{tcolorbox}

\begin{tcolorbox}[colback=white!5!white,colframe=black!75!black,boxrule=1pt,arc=.3em,boxsep=-1mm,fontupper=\footnotesize]
\textbf{\textit{Prompt 4}}
\vspace{3px}

Imagine yourself in the following situation: [From January to March 2030, a new flu strain, NEW FLU, emerged in this country, leading to the first reported cases and the World Health Organisation (WHO) declaring a pandemic.].

\vspace{3px}

Personal Risk: Information provided by public health authorities at this time suggest that the mortality rate is around 1\%. Almost all people who die of the disease are over the age of 75. Young people almost never die from the disease, but they appear to be able to transmit the disease to others. For younger people, symptomatic disease results in an influenza-like illness syndrome lasting around 1 week, with a quarantine period of around 2 weeks, where an infected person is not allowed to go outside their house. It also appears that at least 50\% of people who are infected do not get any symptoms and do not know they are infected.

\vspace{3px}
Your background and personal circumstances are as follows: 
[You are under [AGE] years old, [GENDER] of [RACE] ethnicity living in [CITY]. [CITY CONTEXT]. You're living in a diverse country with varying access to healthcare, differing levels of trust in government and medical institutions, and socioeconomic disparities. 
 
Your household income is [INCOME]. Your education level is [EDUCATION]].
\vspace{3px}

Please use this persona to answer the question below:

\vspace{3px}

\textit{How likely are you to get tested and self-report your symptoms into the new flu public health system if you experience signs of a new flu, which may make you need to self-quarantine?}

\vspace{3px}

In this context, please think step by step before answering Yes or No based on your persona. 
Answer: [Yes or No]

SHORT REASON:   [Explain to me the rationale behind why you made this] decision. Also, Reporting rate: [0-100\% based on the persona]
And Confidence Level: [(Very Certain, Somewhat Certain, Uncertain)]

\vspace{3px}

\textbf{Response Format: }

1. Yes or No Answer

2. Confidence Level: (Very Certain, Somewhat Certain, Uncertain) 

3. Reporting rate: [0-100\% based on the persona]

4. Brief Reason: [one sentence, explain to me the rationale behind why you made this decision.]

\end{tcolorbox}

\begin{tcolorbox}[colback=white!5!white,colframe=black!75!black,boxrule=1pt,arc=.3em,boxsep=-1mm,fontupper=\footnotesize]
\textbf{\textit{Prompt 5}}
\vspace{3px}

Imagine yourself in the following situation: [From January to March 2030, a new flu strain, NEW FLU, emerged in this country, leading to the first reported cases and the World Health Organisation (WHO) declaring a pandemic.].
\end{tcolorbox}
\pagebreak

\begin{tcolorbox}[colback=white!5!white,colframe=black!75!black,boxrule=1pt,arc=.3em,boxsep=-1mm,fontupper=\footnotesize]
\textbf{\textit{Prompt 5 cont.}}
\vspace{3px}



Personal Risk: Information provided by public health authorities at this time suggest that the mortality rate is around 1\%. Almost all people who die of the disease are over the age of 75 years. Young people almost never die from the disease, but they appear to be able to transmit the disease to others. For younger people symptomatic disease results in an influenza-like-illness syndrome lasting around 1 week, with a quarantine period of around 2 weeks where an infected person is not allowed to go outside their house. It also appears that at least 50\% of people who are infected do not get any symptoms and do not know they are infected.

\vspace{3px}
Your background and personal circumstances are as follows: 
[You are under [AGE] years old, [GENDER] of [RACE] ethnicity living in [CITY]. [CITY CONTEXT]. You're living in a diverse country with varying access to healthcare, differing levels of trust in government and medical institutions, and socioeconomic disparities. 
 
Your household income is [INCOME]. Your education level is [EDUCATION]].
\vspace{3px}

In your household, another family member recently experienced similar symptoms and decided to [report/not report] their illness to public health authorities.

\vspace{3px}

Please use this persona to answer the question below:

\vspace{3px}

\textit{How likely are you to get tested and self-report your symptoms into the new flu public health system if you experience signs of a new flu, which may make you need to self-quarantine?}

\vspace{3px}

In this context, please think step by step before answering Yes or No based on your persona. 
Answer: [Yes or No]

SHORT REASON:   [Explain to me the rationale behind why you made this] decision. Also, Reporting rate: [0-100\% based on the persona]
And Confidence Level: [(Very Certain, Somewhat Certain, Uncertain)]

\vspace{3px}

\textbf{Response Format: }

1. Yes or No Answer

2. Confidence Level: (Very Certain, Somewhat Certain, Uncertain) 

3. Reporting rate: [0-100\% based on the persona]

4. Brief Reason: [one sentence, explain to me the rationale behind why you made this decision.]

\end{tcolorbox}

\subsection{Demographic Reporting Rates}\label{appendix:ci-table}

Table \ref{tab:ci-prompt} shows mean reporting rates and 95\% CIs aggregated across five prompt variants. The wide CIs are driven by Prompt 5 (modeling household influence via pre-generated decision banks), which consistently reduces the number of cases reported compared to the other prompts. Despite this aggregate variance, rank orders within income and education remain consistent across cities, confirming the robust demographic gradients from Section \ref{tab:anova}.

\begin{table}[h!]
\centering
\small
\caption{Mean reporting rate and 95\% CI (across five prompt variants, by demographic group and city.}
\label{tab:ci-prompt}
\begin{tabular}{llcc}
\toprule
\textbf{Factor} & \textbf{Group} & \textbf{Atlanta} & \textbf{San Francisco} \\
\midrule
\multirow{2}{*}{Income}
  & Above \$70{,}000          & 82.1\% ($\pm$21.0\%)  & 86.8\% ($\pm$15.2\%)  \\
  & Below \$70{,}000          & 55.8\% ($\pm$14.6\%)  & 65.0\% ($\pm$19.0\%)  \\
\midrule
\multirow{3}{*}{Education}
  & Bachelor or more          & 80.4\% ($\pm$19.7\%)  & 85.9\% ($\pm$17.6\%)  \\
  & Some college              & 70.8\% ($\pm$16.4\%)  & 79.7\% ($\pm$18.2\%)  \\
  & High school or less       & 53.7\% ($\pm$16.8\%)  & 60.6\% ($\pm$16.2\%)  \\
\midrule
\multirow{2}{*}{Age}
  & Over 50                   & 67.1\% ($\pm$17.6\%)  & 74.1\% ($\pm$16.0\%)  \\
  & Under 50                  & 71.6\% ($\pm$15.6\%)  & 78.5\% ($\pm$17.6\%)  \\
\midrule
\multirow{3}{*}{Race}
  & Asian                     & 70.4\% ($\pm$16.8\%)  & 76.2\% ($\pm$17.7\%)  \\
  & Black                     & 65.0\% ($\pm$18.6\%)  & 75.6\% ($\pm$16.6\%)  \\
  & White                     & 72.9\% ($\pm$15.8\%)  & 77.3\% ($\pm$15.9\%)  \\
\bottomrule
\end{tabular}
\end{table}

\end{document}